\documentclass[a4paper,twoside]{article}

\usepackage{epsfig}
\usepackage{subcaption}
\usepackage{calc}
\usepackage{amssymb}
\usepackage{amstext}
\usepackage{amsmath}
\usepackage{amsthm}
\usepackage{multicol}
\usepackage{pslatex}
\usepackage{apalike}
\usepackage{algorithm2e}
\usepackage[bottom]{footmisc}
\usepackage{booktabs}
\usepackage{multirow}
\usepackage{hyperref}

\hypersetup{
  colorlinks,
  citecolor=black,
  urlcolor=blue}

\usepackage{SCITEPRESS}     

\newcommand{\etal}{\textit{et al}. }
\newcommand{\ie}{\textit{i}.\textit{e}. }
\newcommand{\eg}{\textit{e}.\textit{g}. }

\begin{document}

\title{End-to-End Chess Recognition}

\author{\authorname{Athanasios Masouris
, Jan C. van Gemert
}
\affiliation{Computer Vision Lab, Delft University of Technology, Delft, The Netherlands}
}

\keywords{chess recognition, chess dataset, computer vision, deep learning}

\abstract{Chess recognition is the task of extracting the chess piece configuration from a chessboard image. Current approaches use a pipeline of separate, independent, modules such as chessboard detection, square localization, and piece classification. Instead, we follow the deep learning philosophy and explore an end-to-end approach to directly predict the configuration from the image, thus avoiding the error accumulation of the sequential approaches and eliminating the need for intermediate annotations. Furthermore, we introduce a new dataset, \textit{Chess Recognition Dataset (ChessReD)}, that consists of 10,800 real photographs and their corresponding  annotations. In contrast to existing datasets that are synthetically rendered and have only limited angles, ChessReD has photographs captured from various angles using smartphone cameras; a sensor choice made to ensure real-world applicability. Our approach in chess recognition on the introduced challenging benchmark dataset outperforms related approaches, successfully recognizing the chess pieces' configuration in 15.26\% of ChessReD's test images. This accuracy may seem low, but it is $\approx$7x better than the current state-of-the-art and reflects the difficulty of the problem.  The code and data are available through: \url{https://github.com/ThanosM97/end-to-end-chess-recognition}.}


\onecolumn \maketitle \normalsize \setcounter{footnote}{0} \vfill


\begin{figure*}[ht]
\centering
\begin{tabular}{cc}
\includegraphics[width=0.45\linewidth]{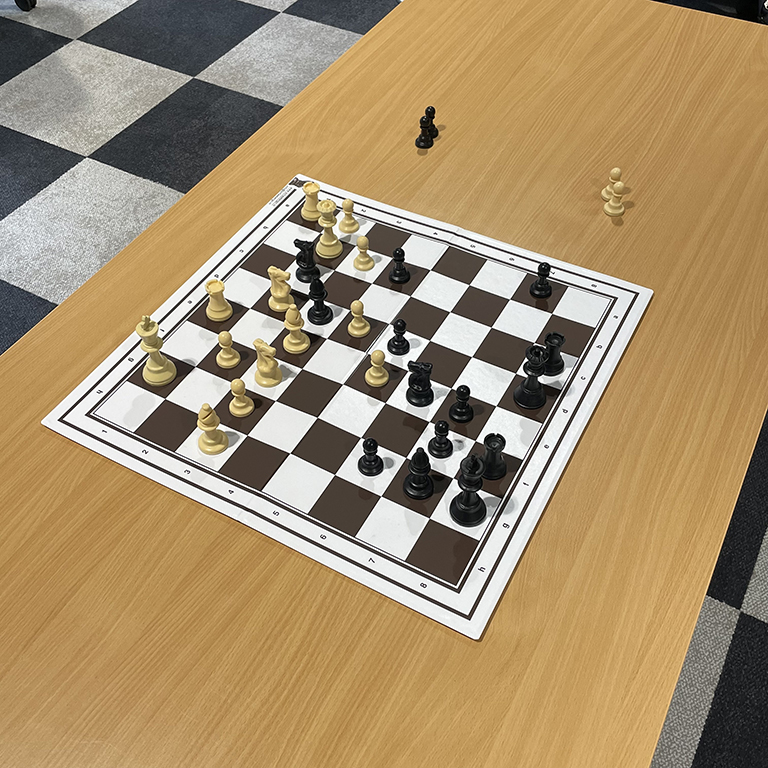}  &
\includegraphics[width=0.45\linewidth]{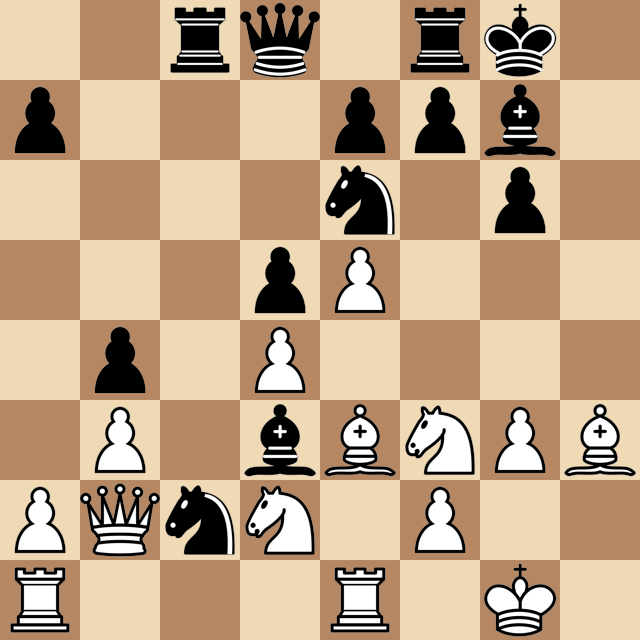}
\end{tabular}
\caption{Chess recognition input image and output configuration.} \label{fig:intro-mot}
\end{figure*}

\section{INTRODUCTION}
\label{sec:intro}
Parsing a chess position from a (smartphone) image eases match analysis and facilitates coaching of children, offering chess positions without writing each move down (Figure \ref{fig:intro-mot}).
Chess position recognition requires accurate identification of each chess piece's type and position on a chessboard configuration. The predominant approach \cite{xie2018chess,mehta2020augmented,czyzewski2020chessboard,wolflein2021determining} is to split it into independent sequential sub-tasks: chessboard detection, square localization, and piece classification within each square. We here depart from the observation that such independent sub-tasks suffer from error accumulation throughout each intermediate independent step. In this paper, we do away with the sequential independent sub-tasks and propose an end-to-end approach for chess recognition, directly predicting the positions of the pieces, with respect to the chessboard, from the entire image.

One key advantage of our approach is that it does not require any human input beyond the input image itself, unlike traditional methods \cite{sokic2008simple,ding2016chessvision,neto2019chess,wolflein2021determining} that rely on user input, such as manually selecting the corners of the chessboard or defining the player's perspective. 
By leveraging a deep neural network, the model is able to extract and use relevant visual features to efficiently recognize the chessboard, predict the pieces' type and positions directly from a single image. 

To evaluate and facilitate our learning-based approach we need a large and good quality dataset. Currently, there is no  real-world chess recognition dataset and thus, we introduce our own novel real-world \textit{Chess Recognition Dataset (ChessReD) }, which we openly share with the community\footnote{\url{https://data.4tu.nl/datasets/99b5c721-280b-450b-b058-b2900b69a90f}
\label{footnote:chessred}}.  Our ChessReD dataset consists of $10,800$ images and their corresponding annotations, allowing us to effectively train and evaluate approaches. Although there might be a human preference for certain viewpoint~\cite{van2011exploiting}, we capture a diverse collection of chessboard images, covering various viewing angles, lighting conditions, camera specifications, and piece configurations. This dataset enables both further research in chess recognition algorithms and a realistic benchmark. 

Our approach in chess recognition on our ChessReD benchmark outperforms related approaches, achieving a board recognition accuracy of 15.26\% ($\approx$7x better than the current state-of-the-art). Our contributions can be summarized as follows. 

\begin{itemize}
    \item We introduce ChessReD: the first dataset of real images for chess recognition, with high viewing angle variability and diversity in chess formations, compared to synthetic alternatives.
    \item We demonstrate end-to-end chess recognition with improved performance compared to alternatives that rely on a sequential pipeline of independent components.
\end{itemize}

\section{RELATED WORK}
\label{sec:related}

\paragraph{Early Approaches in Chess Recognition}
Chess recognition has been a subject of research in the field of computer vision, with several approaches proposed to tackle the challenges associated mainly with detecting the chessboard, but also with recognizing the individual pieces on top of it. Early attempts in the field primarily focused on integrating the chess recognition task as a part of a chess-playing robotic system
\cite{urting2003marineblue,sokic2008simple,matuszek2011gambit,banerjee2011simple}. These systems detected chess moves by comparing the previous frame, with a known game state, to the current frame. They relied on detecting the occupied squares of the chessboard along with the colors of the pieces occupying them. As such, these methods were only able to detect valid chess moves and failed to detect events when two pieces of the same color were swapped, either illegally, or by promoting pawns to another piece type. Additionally, they worked under the assumption that the previous inferred state is correct. Thus, in case of an erroneous move prediction, all of the subsequent game states would be incorrect. Despite the aforementioned issues, the same approach has been adopted by several studies \cite{wang2013chess,koray2016computer,chen2016computer,neto2019chess,chen2019robust,kolosowski2020collaborative} since, with consecutive frames obtained also from a video stream \cite{sokic2008simple,wang2013chess,koray2016computer}. In our paper, contrary to these approaches, we aim to develop a robust method that does not rely on the correctness of the previous inferred state but rather performs chess recognition from a single input image.

\paragraph{Chessboard Detection}
For cases when the previous state is unknown, chess recognition from a single image has also been the focus of studies. Same as with the previously mentioned approaches, the first step is to employ image processing techniques to detect the chessboard and the individual squares; a challenging task even on its own. While it can be simplified by explicitly asking the user to select the four corner points \cite{sokic2008simple,ding2016chessvision,neto2019chess}, modifying the chessboard \cite{urting2003marineblue,banerjee2011simple,Danner2015VisualCR} (\eg using a reference frame around the chessboard), or setting constraints on the camera view angle (\eg top-view) \cite{urting2003marineblue,sokic2008simple,banerjee2011simple,wang2013chess,koray2016computer,neto2019chess,mehta2020augmented}, Neufeld \etal \cite{neufeld2010probabilistic} recognized that these approaches do not represent a general solution, where the chessboard could be in arbitrary locations or the image taken from various camera angles. They proposed a line-based detection method which they combined with probabilistic reasoning. However, they also restricted the setting by expecting the camera angle to be in the range of a human player's perspective. Other studies have also exploited specific viewing angles, such as the players' perspectives \cite{ding2016chessvision,chen2016computer,chen2019robust,wolflein2021determining} or side views \cite{Danner2015VisualCR,quintana2020livechess2fen}. While chessboard detection utilizing the Harris corner detection algorithm \cite{banerjee2011simple,koray2016computer,kolosowski2020collaborative}, template matching \cite{urting2003marineblue,matuszek2011gambit}, or flood fill \cite{wang2013chess} have been explored, in accordance with \cite{neufeld2010probabilistic}, line-based chessboard detection methods have received significant research attention \cite{chen2016computer,chen2019robust,xie2018chess,4700034,xie2018geometry,Danner2015VisualCR,mehta2020augmented}. Czyzewski \etal \cite{czyzewski2020chessboard} introduced an approach based on iterative heat map generation which visualizes the probability of a chessboard being located in a sub-region of the image. After each iteration, the four-sided area of the image containing the highest probability values is cropped and the process is repeated until convergence. While this method involves a great computational overhead, it is able to detect chessboards from images taken from varied angles, with poor quality, and regardless of the state of the actual chessboard (\eg damaged chessboard with deformed edges), with a 99.6\% detection accuracy. W{\"o}lflein and Arandjelovi{\'c} \cite{wolflein2021determining} proposed a chessboard detection method that leveraging the geometric nature of the chessboard, utilizes a RANSAC-based algorithm to iteratively refine the homography matrix and include all the computed intersection points. Their method demonstrated impressive results, since it successfully detected all of the chessboards in their validation dataset. However, it's worth noting that the dataset only included images with viewing angles within the range of a player's perspective. In our paper, we bypass the chessboard detection task, allowing the deep learning models to internally infer its position, and thus we do not rely on user input, or specific viewing angles. 

\paragraph{Piece Classification}
Upon detection of the chessboard, the next step the aforementioned approaches employ is piece classification. A number of techniques have been developed to address this task, either in a 2-way approach (\ie color and type), or 1-way by treating each combination of piece color and type as a separate class (\eg ``white-rook"). In Matuszek \etal \cite{matuszek2011gambit}, the authors utilized one classifier to determine the piece color, and then for each color they trained a type classifier using concatenated scale-invariant feature transform (SIFT) and kernel descriptors for features. A similar approach was used in Ding \cite{ding2016chessvision}, where the author employed SIFT and histogram of oriented gradients (HOG) as feature descriptors for piece type classification with support vector machine (SVM) classifiers. The color was subsequently detected by comparing the binarized image of the square with that of an empty one. Danner and Kafafy \cite{Danner2015VisualCR} and Xie \etal \cite{xie2018chess} argued that the lack of distinguishable texture in small objects, such as chess pieces, leads to insufficient features obtained using SIFT descriptors. Both studies suggested a template matching approach for piece classification, comparing the contours of the detected pieces with reference templates obtained from various angles. Wei \etal \cite{wei2017chess} proposed an approach to recognize pieces using a depth camera and a volumetric convolutional neural network (CNN). More recent studies \cite{mehta2020augmented,czyzewski2020chessboard,quintana2020livechess2fen,wolflein2021determining} follow the 1-way approach for piece classification. They train CNNs to distinguish between 12 or 13 classes of objects (\ie six piece types in both colors and one for empty squares), obtaining impressive results. In Czyzewski \etal \cite{czyzewski2020chessboard}, they also leverage domain knowledge, to improve piece classification, by utilizing a chess engine to calculate the most probable piece configurations and clustering similar figures into groups to deduce formations based on cardinalities. Additionally, given the variation in appearance between chess sets, W{\"o}lflein and Arandjelovi{\'c} \cite{wolflein2021determining} proposed a novel fine-tuning process for their piece classifier to unseen chess sets. In our paper, same as with the chessboard detection task, the classification of the pieces is performed by the deep learning model, without the need to train a separate piece classifier. 

\paragraph{Chess Datasets}
A common problem frequently mentioned in literature \cite{ding2016chessvision,mehta2020augmented,czyzewski2020chessboard,wolflein2021determining} is the lack of a comprehensive chess dataset. This issue hinders not only the ability to fairly evaluate the proposed methods in a common setting but also impedes the deployment of deep learning end-to-end approaches that require a vast amount of data. One proposed solution to this problem is the use of synthetic generated data. In \cite{wei2017chess}, the authors produce point cloud data using a 3D computer-aided design (CAD) model, while Blender \cite{blender} was used to produce synthetic image datasets from a top view camera angle \cite{neto2019chess}, or the player's perspective \cite{wolflein2021determining}.  In our paper, we introduce the first chess recognition dataset of real images, without setting any of the aforementioned restrictions regarding the viewing angles.

\section{CHESS RECOGNITION DATASET (ChessReD)}
\label{sec:data}
The availability of large-scale annotated datasets is critical to the advancement of computer vision research. In this section, we tackle a main issue in the field of chess recognition (\ie the lack of a comprehensive dataset) by presenting a novel dataset\footref{footnote:chessred} specifically designed for this task. The dataset comprises a diverse collection of photographs of chess formations captured using smartphone cameras; a sensor choice made to ensure real-world applicability. 

\paragraph{Data Collection and Annotation}
\label{subsec:data-collection}
The dataset was collected by capturing photographs of chessboards with various chess piece configurations. To guarantee the variability of those configurations, we relied upon the chess opening theory. The Encyclopedia of Chess Openings (ECO) classifies opening sequences into five volumes with 100 subcategories each that are uniquely identified by an ECO code. We randomly selected 20 ECO codes from each volume. Subsequently, each code of this set was randomly matched to an already played chess game that followed the particular opening sequence denoted by the ECO code; thus creating a set of 100 chess games. Finally, using the move-by-move information provided by Portable Game Notations (PGNs) that are used to record chess games, the selected games were played out on a single physical chessboard, commonly used at chess clubs, with images being captured after each move. 

\begin{figure}[ht]
  \begin{subfigure}[t]{.48\columnwidth}
    \centering
    \includegraphics[width=\linewidth]{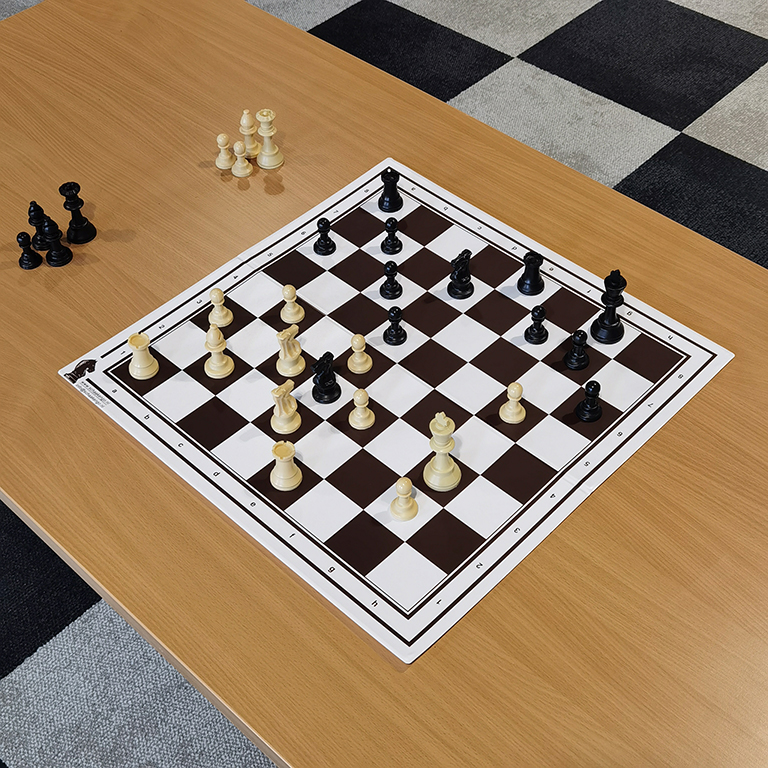}
    \caption{Corner view} \label{subfig:corner-view}
  \end{subfigure}
  \hfill
  \begin{subfigure}[t]{.48\columnwidth}
    \centering
    \includegraphics[width=\linewidth]{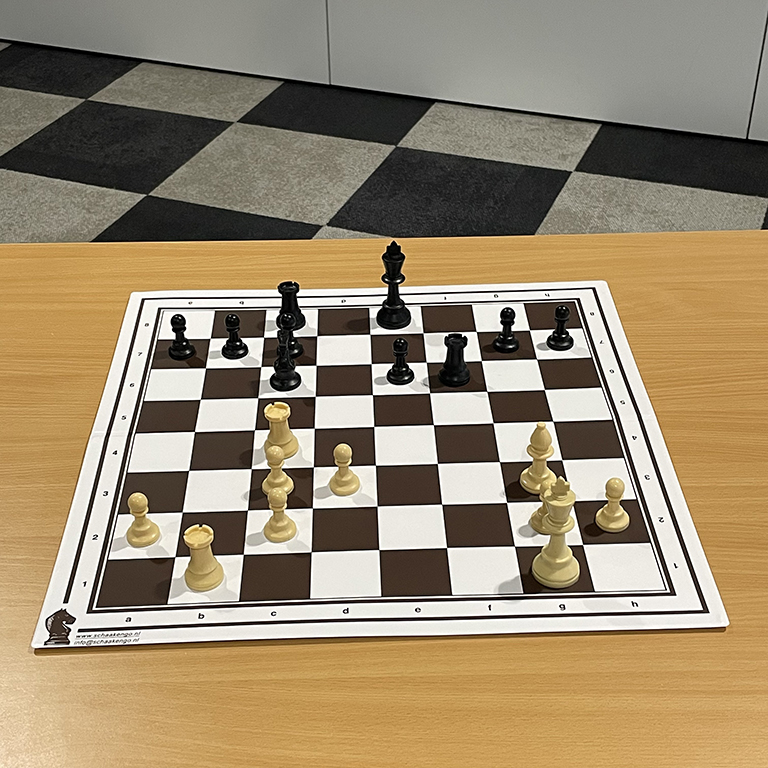}
    \caption{Player view} \label{subfig:player-view}
  \end{subfigure}

  \medskip

  \begin{subfigure}[t]{.48\columnwidth}
    \centering
    \includegraphics[width=\linewidth]{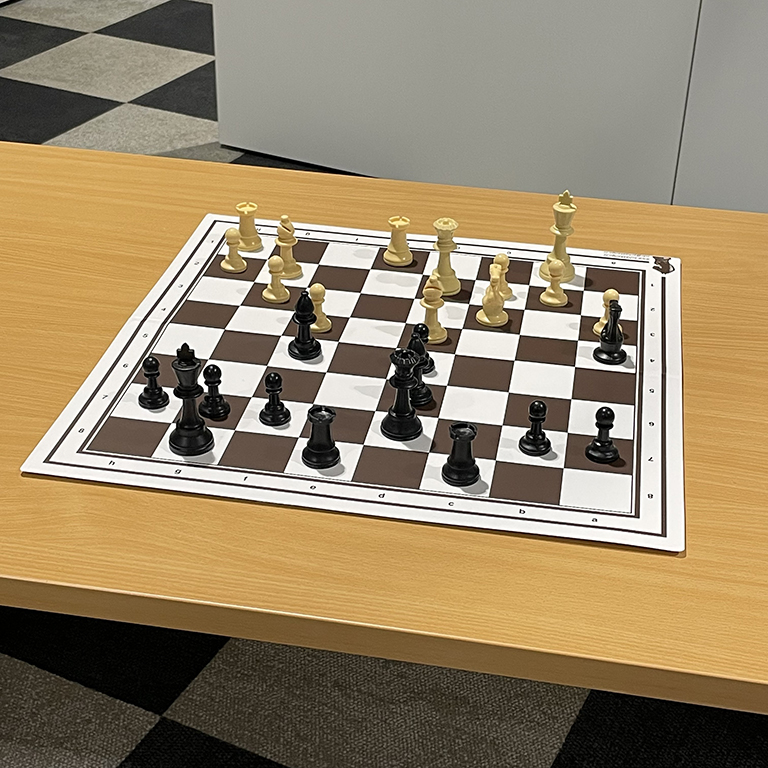}
    \caption{Low angle} \label{subfig:low-angle}
  \end{subfigure}
  \hfill
  \begin{subfigure}[t]{.48\columnwidth}
    \centering
    \includegraphics[width=\linewidth]{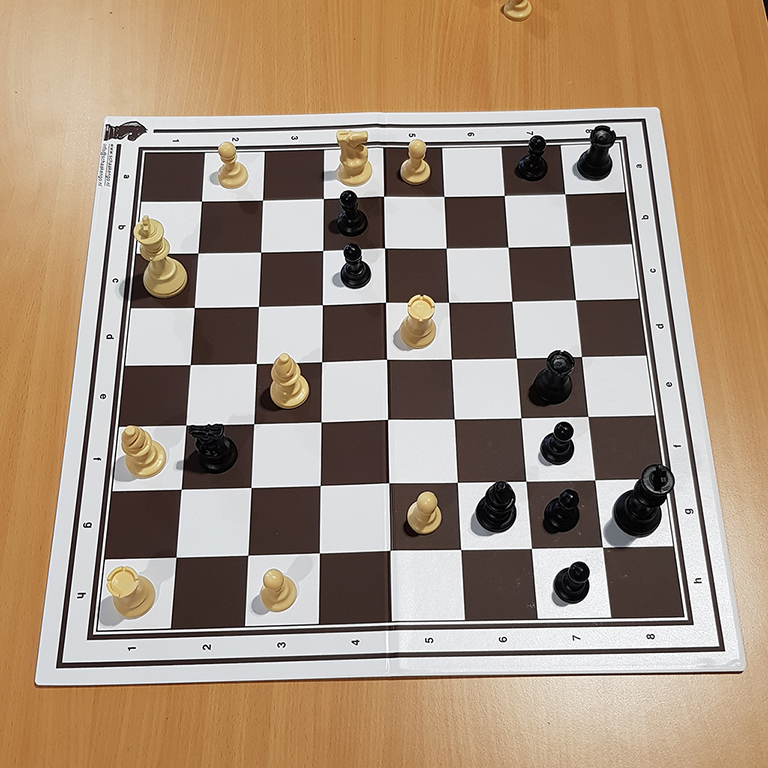}
    \caption{Top view} \label{subfig:top-view}
  \end{subfigure}
  \caption{Image samples from ChessReD.} \label{fig:img-samples}
\end{figure}

Three distinct smartphone models were used to capture the images. Each model has different camera specifications, such as resolution and sensor type, that introduce further variability in the dataset. The images were also taken from diverse angles, ranging from top-view to oblique angles, and from different perspectives (\eg white player perspective, side view, etc.). These conditions simulate real-world scenarios where chessboards can be captured from a bystander's arbitrary point of view. Additionally, the dataset includes images captured under different lighting conditions, with both natural and artificial light sources introducing these variations. Most of these variations are illustrated in the four image samples of Figure \ref{fig:img-samples}. Each of those samples highlights a different challenge in chess recognition. Occlusions between pieces occur more often in images captured from a low angle (Figure \ref{subfig:low-angle}) or a player's perspective (Figure \ref{subfig:player-view}), while pieces are rarely occluded in top-view images (Figure \ref{subfig:top-view}). However, distinct characteristics of pieces (\eg the queen's crown) that could aid the chess recognition task are less distinguishable in a top-view. More samples of ChessReD can be seen in the Appendix.

\begin{figure}
    \centering
    \includegraphics[width=\columnwidth]{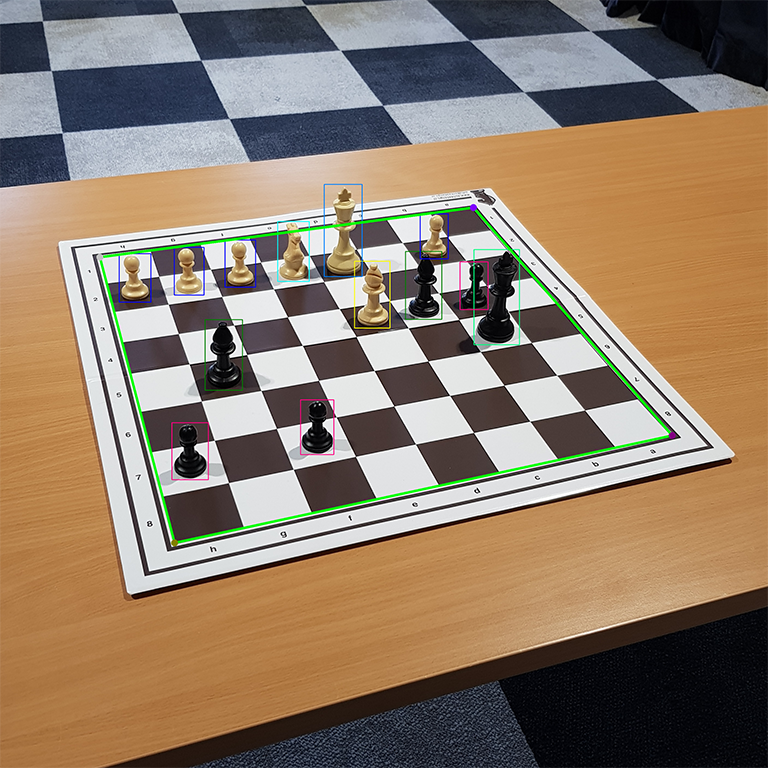}
    \caption{Bounding box and corner point annotations in ChessReD2K.}
    \label{fig:bbox-annotations}
\end{figure}

The dataset is accompanied by detailed annotations providing information about the chess pieces formation in the images. Therefore, the number of annotations for each image depends on the number of chess pieces depicted in it. There are 12 category ids in total (\ie 6 piece types per color) and the chessboard coordinates are in the form of algebraic notation strings (\eg ``a8"). These annotations were automatically extracted from  Forsyth-Edwards Notations (FENs) that were available to us by the games' PGNs. Each FEN string describes the state of the chessboard after each move using algebraic notation for the piece types (\eg ``N" is knight) , capitalization for the piece colors (\ie white pieces are denoted with uppercase letters, while black pieces with lowercase letters), and digits to denote the number of empty squares. Thus, by matching the captured images to the corresponding FENs, the state of the chessboard in each image was already known and annotations could be extracted. To further facilitate research in the chess recognition domain, we also provide bounding-box and chessboard corner annotations for a subset of 20 chess games. An annotated sample is presented in Figure \ref{fig:bbox-annotations}. The different colors for the corner points represent the four distinct corner annotations (\ie bottom-left, bottom-right, top-left, and top-right) that are relative to the white player's perspective. For instance, the corner annotated with the red color in Figure \ref{fig:bbox-annotations} is a \textit{bottom-left} corner. The discrimination between these different types of corners provides information about the orientation of the chessboard that can be leveraged to determine the image's perspective and viewing angle.

\paragraph{Data Statistics}
\label{subsec:data-stats}
The dataset consists of 100 chess games, each with an arbitrary number of moves and therefore images, amounting to a total of 10,800 images being collected. The dataset was split into training, validation, and test sets following an 60/20/20 split. Since two consecutive images of a chess game differ only by one move, the split was performed on game-level to ensure that quite similar images would not end up in different sets. The split was also stratified over the three distinct smartphone cameras (Apple iPhone 12, Huawei P40 pro, Samsung Galaxy S8) that were used to capture the images. Table \ref{tab:stats-images} presents an overview of the image statistics per smartphone. The three smartphone cameras introduced variations to the dataset based on the distinct characteristics of their sensors. For instance, while the image resolution for the Huawei phone was 3072x3072, the resolution for the remaining two models was 3024x3024.

\begin{table}[ht]
\centering
\caption{Overview of the image statistics.}
\label{tab:stats-images}
\begin{tabular}{cccc}
\toprule
\multicolumn{1}{c}{\multirow{2}{*}{\textbf{Smartphone}}} & \multicolumn{3}{c}{\textbf{Number   of images}} \\
\multicolumn{1}{c}{}                                              & \textbf{Train}  & \textbf{Val}  & \textbf{Test} \\
\midrule
Apple iPhone 12                                                   & 2,146            & 851           & 638           \\
Huawei P40 pro                                                    & 2,102            & 638           & 871           \\
Samsung Galaxy S8                                                 & 2,231            & 703           & 620           \\
\textbf{Total}                                                    & \textbf{6,479}   & \textbf{2,192} & \textbf{2,129} \\
\bottomrule
\end{tabular}

\end{table}

Table \ref{tab:stats-anns} presents an overview of the annotations in ChessReD. In Table \ref{tab:stats-anns-pos} illustrates a significant imbalance between  annotations for the piece type ``Pawn" and the rest of the pieces. This was to be expected since every chess game starts with 8 pawns in each side and only one or two of the remaining piece types. Regarding the colors of the pieces, no imbalance is detected in the dataset. Additionally, while annotations about the position of the pieces in algebraic notation are available for every image in the dataset, we provide bounding box and chessboard corner annotations only for a subset of 20 randomly selected games (2,078 images) from the train, validation, and test sets. For this subset we followed a 70/15/15 split stratified over the smartphone cameras, which led to a total of 14 training games (1442 images), 3 validation games (330 images), and 3 test games (306 images) being annotated. In Table \ref{tab:stats-anns-bboxes} we can see an overview of the annotation statistics for this subset, named ChessReD2K.

\begin{table*}[t]
\centering
\caption{Overview of the annotation statistics.} \label{tab:stats-anns}
\begin{minipage}{0.48\textwidth}
    \centering
    \subcaption{Piece positions on the chessboard in ChessReD}\label{tab:stats-anns-pos}
    \resizebox{\textwidth}{45pt}{
    \begin{tabular}{ccccccc}
    \toprule
    \multirow{3}{*}{\textbf{Piece type}} & \multicolumn{6}{c}{\textbf{Number of instances}}                                                          \\
                                         & \multicolumn{2}{c}{\textbf{Train}} & \multicolumn{2}{c}{\textbf{Val}} & \multicolumn{2}{c}{\textbf{Test}} \\
                                         & \textbf{Black}   & \textbf{White}  & \textbf{Black}  & \textbf{White} & \textbf{Black}  & \textbf{White}  \\
    \midrule
    Pawn                                 & 35,888            & 35,021           & 11,410           & 11,042          & 11,616           & 11,472           \\
    Rook                                 & 9,317             & 9,260            & 2,605            & 2,876           & 2,992            & 3,077            \\
    Knight                               & 6,158             & 6,471            & 2,222            & 2,206           & 2,032            & 2,202            \\
    Bishop                               & 6,681             & 6,768            & 2,167            & 2,003           & 2,301            & 2,067            \\
    Queen                                & 4,076             & 3,996            & 1,011            & 1,013           & 1,145            & 1,109            \\
    King                                 & 6,479             & 6,479            & 2,192            & 2,192           & 2,129            & 2,129           \\
    \bottomrule
    \end{tabular}
    }
\end{minipage}%
\hfill
\begin{minipage}{0.48\textwidth}
    \centering
    \subcaption{Bounding boxes in ChessReD2K}\label{tab:stats-anns-bboxes}
    \resizebox{\textwidth}{45pt}{
    \begin{tabular}{ccccccc}
    \toprule
    \multirow{3}{*}{\textbf{Piece type}} & \multicolumn{6}{c}{\textbf{Number of instances}}                                                          \\
                                         & \multicolumn{2}{c}{\textbf{Train}} & \multicolumn{2}{c}{\textbf{Val}} & \multicolumn{2}{c}{\textbf{Test}} \\
                                         & \textbf{Black}   & \textbf{White}  & \textbf{Black}  & \textbf{White} & \textbf{Black}  & \textbf{White}  \\ 
    \midrule
    Pawn                                 & 8,059             & 7,653            & 1,511            & 1,625           & 1,719            & 1,624            \\
    Rook                                 & 2,293             & 2,250            & 471              & 447             & 433             & 433             \\
    Knight                               & 1,276             & 1,423            & 178              & 274             & 278             & 278             \\
    Bishop                               & 1,578             & 1,607            & 380              & 335             & 296             & 304             \\
    Queen                                & 862               & 838              & 125              & 126             & 157             & 160             \\
    King                                 & 1,442             & 1,442            & 330              & 330             & 306             & 306 \\
    \bottomrule
    \end{tabular}
    }
\end{minipage}
\end{table*}

\section{METHOD: END-TO-END CHESS RECOGNITION}
Unlike the conventional pipeline in chess recognition that involves separate, independent, modules of chessboard detection, square localization, and chess piece classification, the focus of this study was to explore an end-to-end approach that tackles the recognition task utilizing only a single image as input. Thus, the developed method should take as input an image of a chessboard and output the type and the positions of the pieces relative to the board. To this end, we experimented with two different solutions by treating the problem either as a \textit{multi-class multi-label classification} or as a \textit{relative object detection} task.

\paragraph{Classification approach}
Here, each chessboard square is a distinct label. Since there are 64 squares in each image, and thus 64 labels, this problem is treated as a multi-label classification task. Each square in the chessboard is either unoccupied or occupied by one of the 12 different types of pieces (\ie 6 per color) in chess. Therefore, to each label we assign one of 13 classes (\ie 12 piece types and 'empty'). By formulating it as multi-label multi-class classification, the goal is for the model to learn the intricate relationships and visual patterns associated with the individual squares.

\paragraph{Relative Object Detection Approach}
\label{subsec:approach:relative-detection}
In addition to the multi-class multi-label classification approach, we explore a novel technique for chess recognition which we call \textit{relative object detection}. Contrary to conventional object detection methods that predict bounding box coordinates in terms of absolute position in the image frame, our modified method aims to predict the x and y coordinates of the objects relative to the chessboard grid in the image. In this manner, discrete coordinates that align with the chessboard positions are used to provide spatial information of its layout. For instance, the relative position (0,0) corresponds to the chessboard square denoted by ``a8" in chess algebraic notation. Furthermore, since we only need to predict the relative coordinates, we can omit the height and width estimation, effectively bypassing the complexities of the bounding box size estimation.

\subsection{Implementation Details}

\paragraph{Classification}
\label{subsec:training:classification}
For the classification approach, we employed a ResNeXt \cite{xie2017aggregated} model. Their introduced concept of ``cardinality" (\ie the number of parallel branches used in each residual block) both enables deeper architectures with reduced computation complexity and allows complex representations to be learned by aggregating the information of the multiple branches. Because of these modifications, this family of models can achieve impressive results in image classification.

For our experiments, we trained the \textit{resnext101\_32d} variant, which uses a cardinality of 32 and a width (\ie number of filters) of 8. This means that each residual block in the network contains 32 parallel convolutional layers, and each of them has a width of 8. Additionally, there are 101 layers in the network, which amount to a total of 88.8M parameters. We trained this model from scratch for 200 epochs, with early stopping enabled and a batch size of 8 samples, using a cross-entropy loss function. We used an Adam \cite{kingma2014adam} optimizer with a learning rate of 0.001, which was reduced to 0.0001 after the 100th epoch. The same training recipe was used to train ResNext on all of the datasets in our experiments (Section \ref{sec:expr}).

\paragraph{Relative Object Detection}
As mentioned in Section \ref{subsec:approach:relative-detection}, the goal is to predict a set of coordinates x and y for the chess pieces relative to the chessboard grid in the image. Thus, traditional object detection models that use Region Proposal Networks (RPNs) \cite{ren2015faster} or anchor boxes \cite{redmon2016you} are not suitable for this task, since they output absolute image coordinates. However, a single end-to-end object detection model, like Detection Transformer (DETR) \cite{carion2020end}, that directly predicts bounding boxes and class labels for objects in an image could be employed. DETR uses a transformer encoder-decoder architecture, with the encoder taking as input a feature map produced by a convolutional backbone network and the decoder generating the final predictions using self-attention mechanisms to attend to different parts of the feature map. 

For our experiments, we attempted to train a modified version of DETR that predicts relative object coordinates and omits the height and width dimensions for the bounding boxes of the traditional object detection task. ResNext101\_32d was used as a backbone network for feature extraction. We set the number of queries (\ie the maximum number of objects that DETR can detect in an image) to 32, since each chessboard can have at most 32 chess pieces on top of it. DETR also requires a separate class for ``background", which in our case corresponds to ``empty" squares. Thus, the number of classes that the model is trained to predict is 13 (\ie 12 piece types and background). We trained the model from scratch for a total of 800 epochs, with early stopping enabled and a batch size of 8 samples, using DETR's default bipartite matching loss for set predictions, which takes into account both the class prediction and the similarity of the predicted and ground truth coordinates. We used an AdamW \cite{loshchilov2017decoupled} optimizer with separate learning rates for the backbone network and the encoder-decoder architecture. In particular, the initial learning rates were set to $10^{-5}$ and $10^{-6}$ for the encoder-decoder and backbone, respectively, and a scheduler was used to reduce both by a factor of 10 every 300 epochs. Furthermore, gradient clipping was used with a threshold of $0.1$.

However, the training of this modified DETR variant for chess recognition did not yield optimal results, with the model being unable to successfully detect chess pieces in the images of ChessReD. This issue could potentially be linked to DETR's inherent limitation in detecting small objects \cite{carion2020end,zhu2020deformable}, especially when considering the intricacies of the dataset (\eg occlusions) and the relatively small sizes of individual pieces. Due to the unsuccessful convergence of the DETR variant, it will not be used in the experiments of Section \ref{sec:expr}. Nevertheless, end-to-end relative object detection with transformers is a promising area that should be further investigated, with the focus being on refining the model architecture (\cite{zhu2020deformable}) or the training objective.

\section{EXPERIMENTS}
\label{sec:expr}

\subsection{Exp1: Comparison with the State-of-the-art}
To the best of our knowledge, the current state-of-the-art approach in chess recognition, namely \textit{Chesscog}, was introduced in W{\"o}lflein and Arandjelovi{\'c}~\cite{wolflein2021determining}.  In their experiments, Chesscog achieved a 93.86\% accuracy in chess recognition on a synthetic dataset~\cite{wolflein2021dataset} rendered in Blender\cite{blender}, with a 0.23\% per-square error rate. Additionally, the authors introduced a few-shot transfer learning approach to unseen chess sets and the system demonstrated a 88.89\% accuracy and 0.17\% per-square error rate, when tested on a set of previously unseen images of chessboards. In this section, we will compare the performance of our approach with that of Chesscog's, both on their Blender dataset and on our newly introduced ChessReD.

\subsubsection{Current SOTA: Chesscog}
Chesscog~\cite{wolflein2021determining} does chess recognition using a pipeline that involves chessboard detection, square localization, occupancy classification, and piece classification. It uses the geometric nature of the chessboard to detect lines and employs a RANSAC-based algorithm to compute a projective transformation of the board onto a regular grid. Subsequently, individual squares are localized based on the intersection points and an occupancy classifier is used on each individual square. Finally, the pieces on the occupied squares are classified into one of 12 classes, using a pre-trained piece classifier. The piece classifier is used on image patches of the squares that are heuristically cropped by extending the bounding boxes based on the square's location on the chessboard. During inference, the user must manually input the specific player's perspective (\ie ``white" or ``black") to determine the orientation of the board.

\subsubsection{Comparison to Chesscog on their Synthetic Blender Dataset}
First, we compare and evaluate the performance of our classification approach on Chesscog's synthetic Blender dataset~\cite{wolflein2021determining}. The Blender dataset comprises a set of 4,888 synthetic chessboard images with distinct piece configurations, multiple lighting conditions, a limited range of viewing angles (between 45° and 60° to the board), and images taken only from the players' perspectives. We trained our ResNeXt model following the recipe described in Section \ref{subsec:training:classification} on the dataset's training samples. Subsequently, we evaluated our trained model's performance on the test set. The first two columns of Table \ref{tab:evaluation} demonstrate the evaluation results for both approaches on the Blender dataset. We use the same evaluation metrics as in W{\"o}lflein and Arandjelovi{\'c} \cite{wolflein2021determining}.

Chesscog outperforms our classification approach across all metrics. For the percentage of boards with no mistakes, which reveals a model's ability to achieve perfect board recognition, Chesscog demonstrates a significant advantage with 93.86\% of boards correctly predicted, while ResNeXt achieves this only in 39.76\% of the boards. When one mistake is allowed per board prediction, Chesscog can successfully recognize almost all of the boards, with ResNeXt's performance improving significantly and reaching 65.2\%. Chesscog's superiority is also corroborated by the substantially lower mean number of incorrect squares per board (0.15 vs.\ 1.19 for ResNeXt) and per-square error rate (0.23\% vs.\ 1.86\% for ResNeXt).

\begin{table*}[ht]
\centering
\caption{Performance evaluation for Chesscog's and our classification approach's (ResNeXt) predictions on the corresponding test sets. ChessReD* represents the subset of the test images in which Chesscog could detect the chessboard.} \label{tab:evaluation}
\begin{tabular}{ccccccc}
\toprule
                                                & \multicolumn{2}{c}{\textbf{Blender Dataset}} & \multicolumn{2}{c}{\textbf{ChessReD}} & \multicolumn{2}{c}{\textbf{ChessReD*}} \\
\textbf{Metric}                                 & \textbf{Chesscog}     & \textbf{ResNeXt}     & \textbf{Chesscog}    & \textbf{ResNeXt}    & \textbf{Chesscog}     & \textbf{ResNeXt}    \\ \midrule
Mean incorrect squares per board      & \textbf{0.15}         & 1.19                 & 42.87                & \textbf{3.40}       & 12.96                 & \textbf{3.35}       \\
Boards with no mistakes (\%) & \textbf{93.86\%}       & 39.76\%               & 2.30\%                & \textbf{15.26\%}     & 6.69\%                 & \textbf{15.30\%}     \\
Boards with $\leq1$ mistake (\%) & \textbf{99.71\%}       & 65.20\%               & 7.79\%                & \textbf{25.92\%}     & 22.67\%                & \textbf{27.04\%} \\
Per-square error rate (\%)                           & \textbf{0.23\%}        & 1.86\%                & 73.64\%               & \textbf{5.31\%}      & 39.57\%                & \textbf{5.24\%} \\ \bottomrule
\end{tabular}
\end{table*}

\subsubsection{How does the Classification Approach Compare to Chesscog on the Real ChessReD Dataset?}
\label{subsubsec:evaluation:chess-dataset}
In this section, we compare the performance of our approach with that of Chesscog's on our ChessReD. We trained our ResNeXt model, using again the recipe of Section \ref{subsec:training:classification}, and finetuned the Chesscog classifiers as mentioned in \cite{wolflein2021determining}, using two images of the starting position from both players' perspectives. Furthermore, for a fair comparison we needed to take into account that Chesscog cannot infer the orientation of the chessboard and requires for it to be manually inputted. Since this information is not available in our dataset, we address it by generating all possible orientations for the detected chessboards during evaluation.

Both approaches faced increased challenges when tested on ChessReD, resulting in a performance drop across all metrics, as seen in Table \ref{tab:evaluation}. While our ResNeXt model can still demonstrate competitive results, recognizing successfully 15.26\% of boards with no mistakes and 25.92\% of boards with less than one mistake, Chesscog's accuracy decreases significantly, achieving only 2.3\% and 7.79\% in these metrics, respectively. Chesscog's performance deterioration is also evident by its 42.87 incorrect squares per board on average and the 73.65\% per-square error rate. ResNeXt's performance for these metrics was 3.40 and 5.31\%, respectively. 

Upon further investigation, one important factor that led to Chesscog's performance degradation was the inaccurate results of the chessboard detection process and the accumulation of the error throughout the pipeline. While the limited range of angles present in the Blender dataset of the previous section enabled Chesscog to achieve 100\% accuracy in chessboard detection, the corresponding accuracy in our dataset is 34.38\%. This issue highlights the sensitivity of the image processing algorithms employed for chessboard detection to their hyperparameters and the necessity to finetune them across different datasets.

To further compare the performance of both approaches, we conducted the same evaluation without taking into account the failed chessboard detections by Chesscog. In the last two columns of Table \ref{tab:evaluation}, we evaluate the performance of both approaches on the subset of the ChessReD's test set (denoted as ChessReD*) consisting of the 34.38\% (732) of the images in which Chesscog was able to detect the chessboard. While Chesscog's performance shows significant improvement when we don't consider those erroneous chessboard detections, it remains inferior in comparison to the results achieved by our classification approach across all metrics.

\subsection{Ablation: Chessboard Markings}
\begin{figure}[ht]
\centering
  \begin{subfigure}[t]{0.48\columnwidth}
    \centering
    \includegraphics[width=\linewidth]{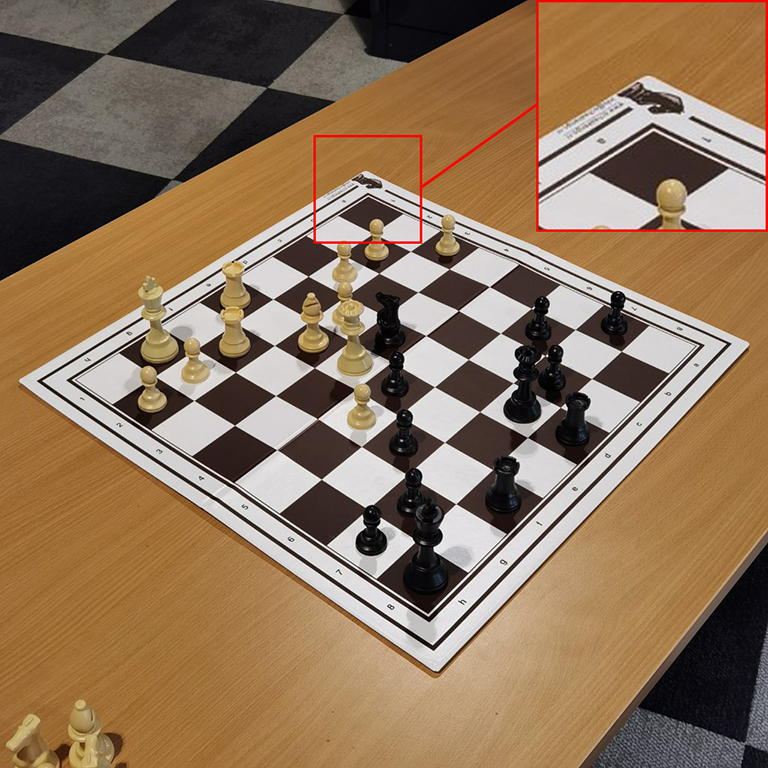}
    \caption{With marks} \label{subfig:ablation}
  \end{subfigure}
  \hfill
  \begin{subfigure}[t]{0.48\columnwidth}
    \centering
    \includegraphics[width=\linewidth]{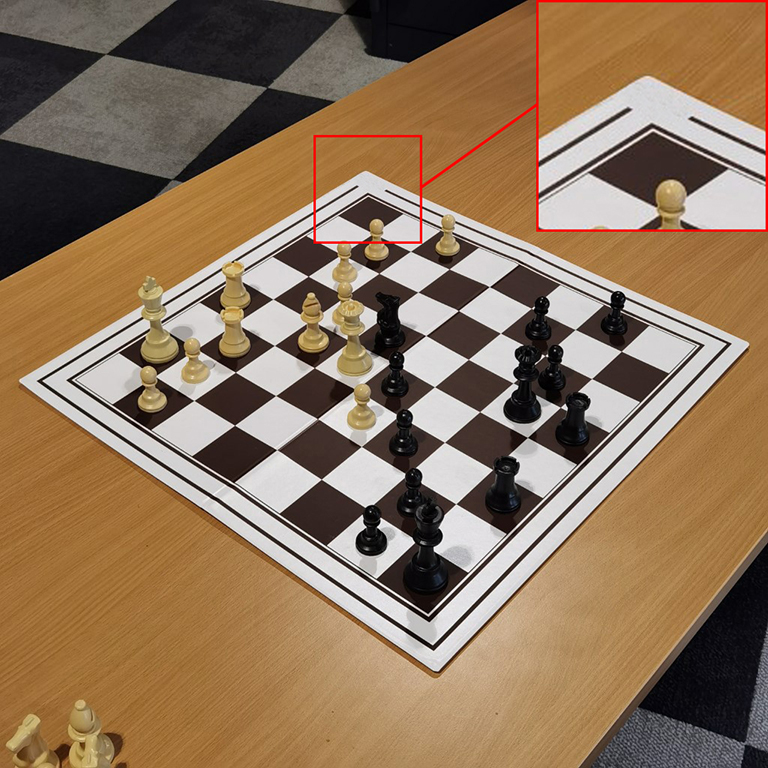}
    \caption{Without marks} \label{subfig:ablation-without}
  \end{subfigure}
\caption{Sample pair of images for the ablation study.} \label{fig:ablation-sample}
\end{figure}

One significant advantage of our approaches is that they do not require any further input to determine the orientation of the chessboard in an image. Yet, the visual cues used by the models to deduce the chessboard's orientation remain unclear. To this end, we conducted an ablation study to investigate whether the ResNeXt model relies on specific marks of the chessboard (\eg bishop logo in Figure \ref{subfig:ablation}) which can, for example be detected~\cite{karaoglu2017text}, to subsequently determine the board's orientation and recognize the chess configuration.

We hypothesize that the necessity of those marks for successful chess recognition increases with the number of moves that have been made prior to capturing the image. The intuition behind this hypothesis is that in the early game of chess, the majority of the pieces remain in their starting position, so determining the boards orientation poses less of a challenge, while in the end game, only a few pieces remain on the board and they are usually far from their starting position. To validate this hypothesis, we created a dataset consisting of 30 test images that were randomly selected from the subset of images that the model was able to successfully recognize in the evaluation of Section \ref{subsubsec:evaluation:chess-dataset}. The test images were evenly distributed across three categories: early-game, mid-game, and end-game. These categories correspond to images that were taken when less than 30, more than 30 but less than 75, or more than 75 moves had been made prior to capturing the images, respectively. Subsequently, we manually removed the marks, such as the bishop logo or the algebraic chess notation on the sides, and evaluated again the performance of ResNeXt on this subset of 30 modified images. A sample pair of images is illustrated in Figure \ref{fig:ablation-sample}. The model achieved an overall accuracy (\ie boards with no mistakes) of 66.6\% on this subset, with a perfect recognition in the early-game images, 60\% accuracy in mid-game images, and 40\% accuracy in end-game images.

\section{DISCUSSION}
The evaluation on the Blender dataset revealed that Chesscog outperforms our classification approach. However, further experimentation on our newly introduced ChessReD showcased a shift in both methods' performances, with ResNeXt surpassing Chesscog across all metrics. It is evident that Chesscog's low chessboard detection rate (34.38\%), which is attributed to the diverse angles and occlusions introduced by our dataset, significantly contributed to that shift, while the specific range of angles used in the Blender dataset enabled Chesscog to successfully detect the chessboard in all cases and achieve a remarkable end-to-end performance.

The ablation study provided significant insights into our ResNeXt model's reliance on specific marks for determining chessboard orientation, and therefore chess recognition. The study confirmed our hypothesis that the necessity of those marks increases with the number of moves made prior to image capture. The model achieved higher accuracy in early-game images, where most of the pieces remained in their starting positions, and lower accuracy in end-game images, where only a few pieces were still on the board and farther from their starting positions. While depending on such marks could be challenging in cases where they are absent or obscured, it could prove to be an advantage in end-game states in which even human annotators can have trouble determining the board's orientation without them.

\subsection{Limitations}
While our study sheds light on the importance of end-to-end deep learning approaches for chess recognition, the limitations of these solutions should also be considered. An inherent weakness of the classification approach is its inability to recognize labels that are absent from the dataset that it was trained on. For instance, if a specific piece/square combination was first seen at inference time, the model would be unable to assign the corresponding label. On the other hand, the relative object detection approach would not encounter this issue, but as a transformer-based solution it's difficult to converge when trained on a small dataset. Finally, finetuning these models on previously unseen data would require considerably more resources and compute time compared to finetuning a simple CNN piece classifier in the sequential approaches. 

Regarding ChessReD, although including a single chess set in the images was a design choice, this lack of diversity impedes the development of solutions with broader applicability. Yet, it is feasible to enhance the dataset by collecting varied data with relative positional annotations (\ie FEN strings instead of bounding boxes) from chess tournaments recordings, where the players are obliged to annotate their every move.

\subsection{Conclusion}
\label{sec:conclusion}
Our experiments demonstrate the effectiveness of our classification approach in chess recognition tasks, while also revealing Chesscog's advantages on certain datasets. However, with the focus being on real-world applicability, the ChessReD dataset, consisting of real images with varied angles and perspectives, poses a more challenging benchmark for chess recognition, and thus the experimental results establish our approach as the state-of-the-art method for this task. Moving forward, improving the model's ability to generalize by either enhancing the dataset, or incorporating domain adaptation techniques, should be explored. Additionally, the relative object detection approach, if converged, may constitute a more robust solution for chess recognition, and thus requires further studying.

\section*{ACKNOWLEDGEMENTS}

This project is supported in part by NWO (project VI.Vidi.192.100).

\bibliographystyle{apalike}
{\small
\bibliography{egbib}}

\section*{APPENDIX}

\begin{figure*}[ht]
  \begin{subfigure}[t]{.66\columnwidth}
    \centering
    \includegraphics[width=\linewidth]{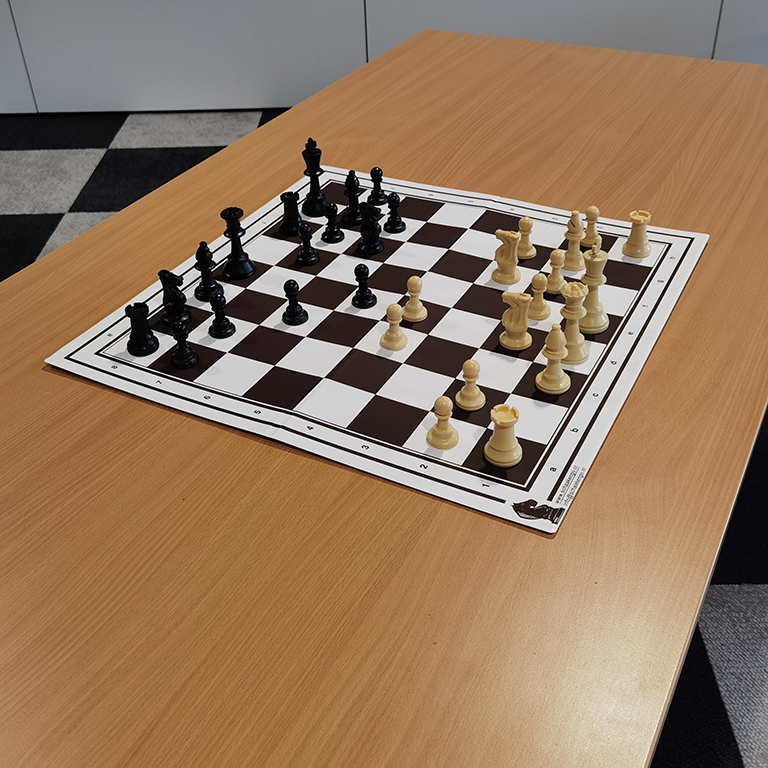}
  \end{subfigure}
  \hfill
  \begin{subfigure}[t]{.66\columnwidth}
    \centering
    \includegraphics[width=\linewidth]{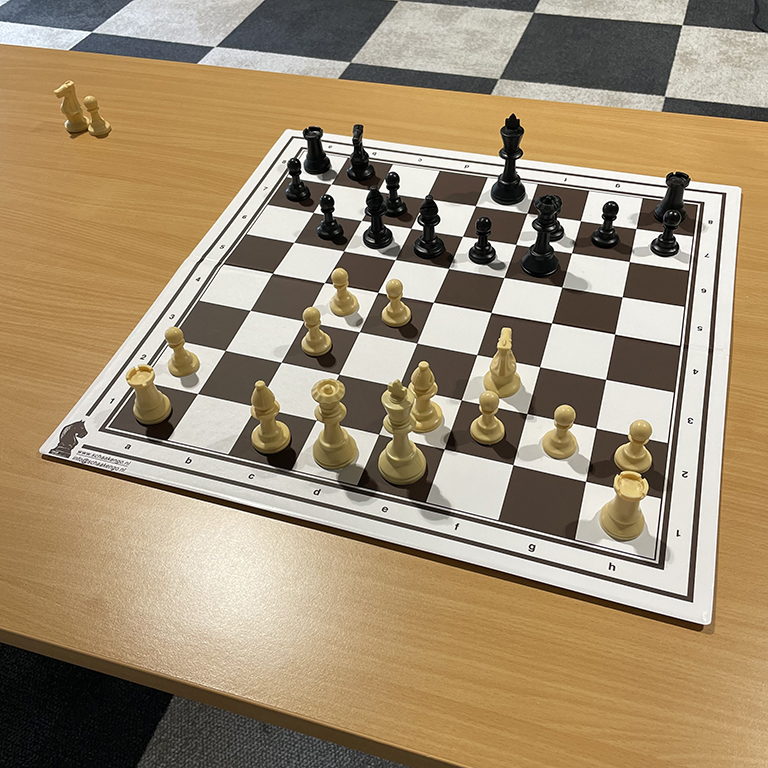}
  \end{subfigure}
  \hfill
  \begin{subfigure}[t]{.66\columnwidth}
    \centering
    \includegraphics[width=\linewidth]{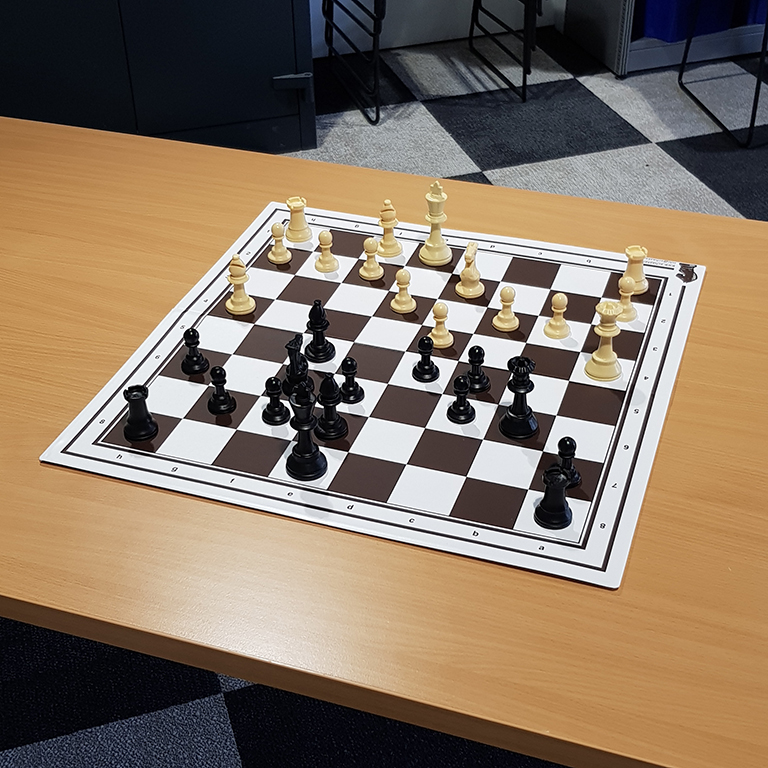}
  \end{subfigure}
  \hfill
  \caption{Early-game (less than 30 moves) samples from ChessReD.} \label{apdx:fig:early-game}
\end{figure*}

\begin{figure*}[ht]
  \begin{subfigure}[t]{.66\columnwidth}
    \centering
    \includegraphics[width=\linewidth]{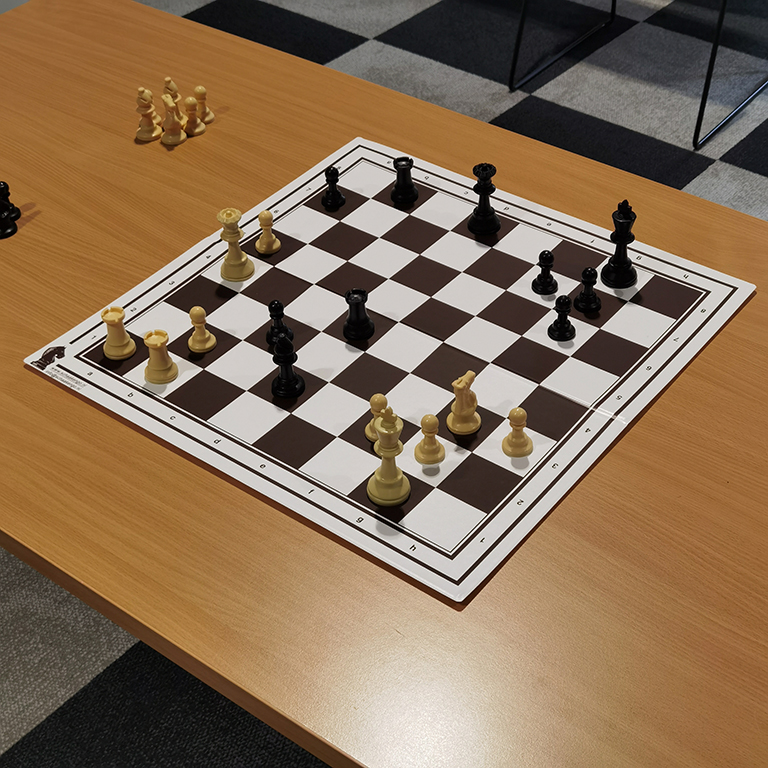}
  \end{subfigure}
  \hfill
  \begin{subfigure}[t]{.66\columnwidth}
    \centering
    \includegraphics[width=\linewidth]{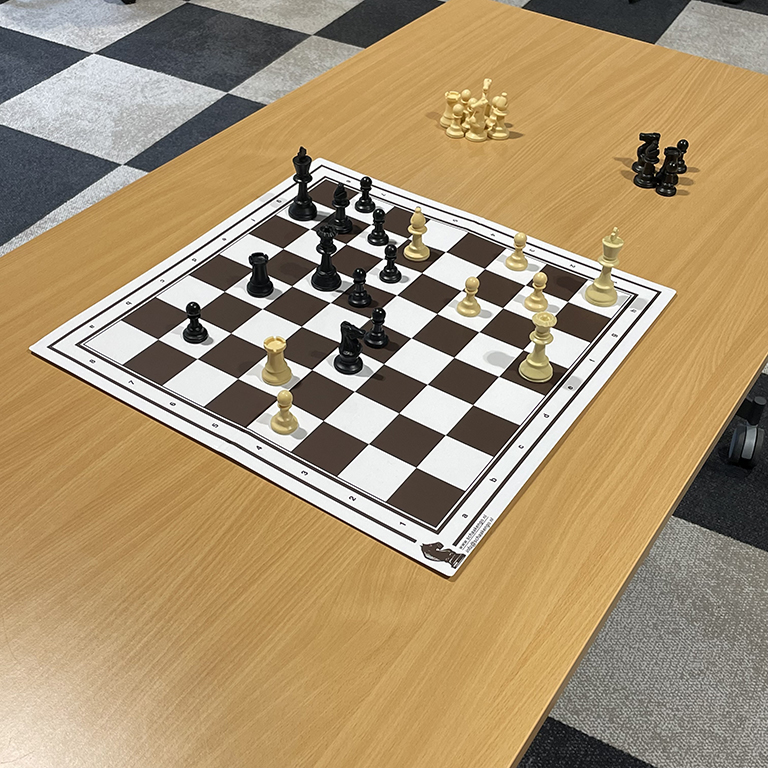}
  \end{subfigure}
  \hfill
  \begin{subfigure}[t]{.66\columnwidth}
    \centering
    \includegraphics[width=\linewidth]{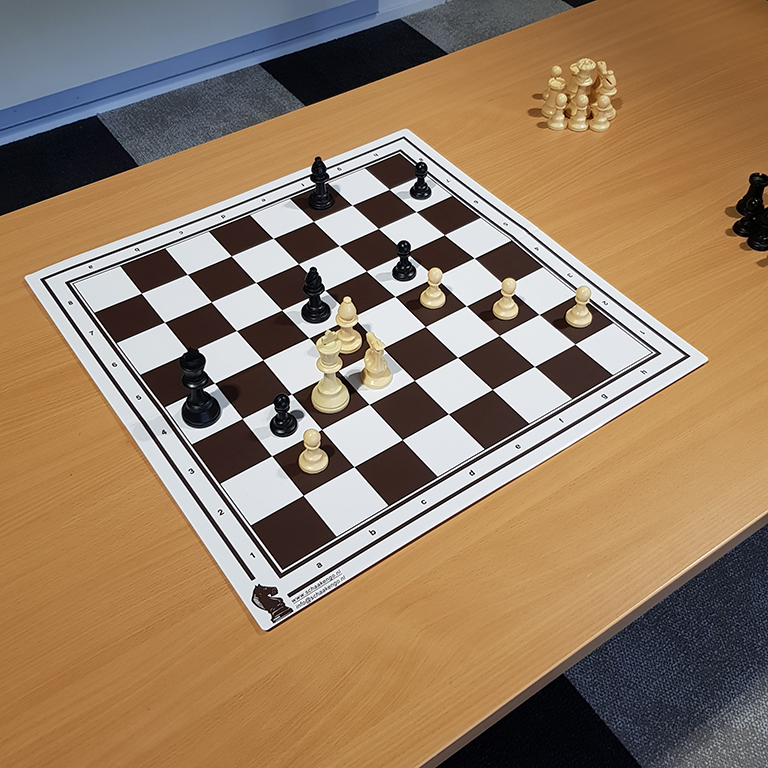}
  \end{subfigure}
  \hfill
  \caption{Mid-game (more than 30 and less than 75 moves) samples from ChessReD.} \label{apdx:fig:mid-game}
\end{figure*}

\begin{figure*}[ht]
  \begin{subfigure}[t]{.66\columnwidth}
    \centering
    \includegraphics[width=\linewidth]{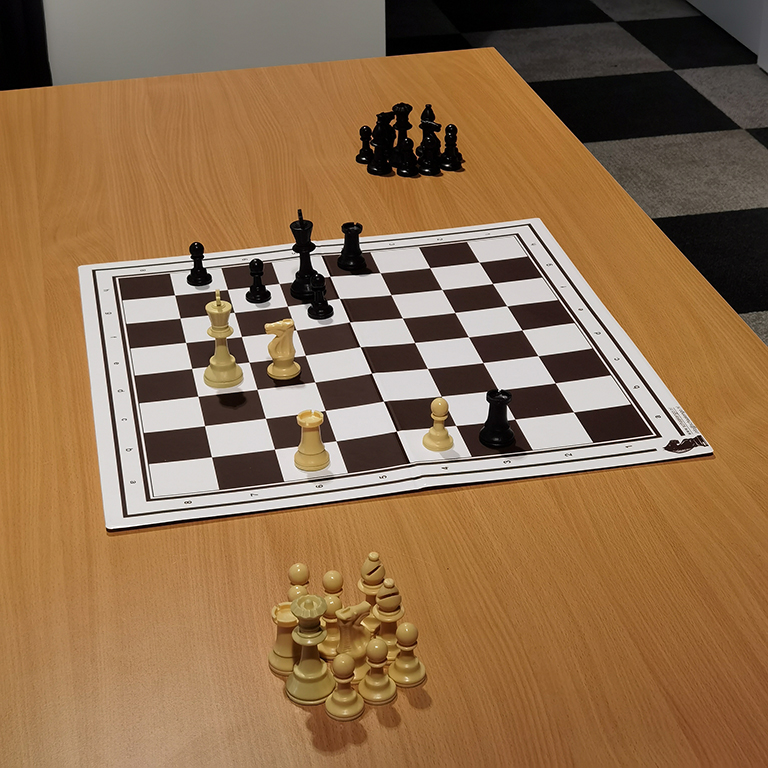}
  \end{subfigure}
  \hfill
  \begin{subfigure}[t]{.66\columnwidth}
    \centering
    \includegraphics[width=\linewidth]{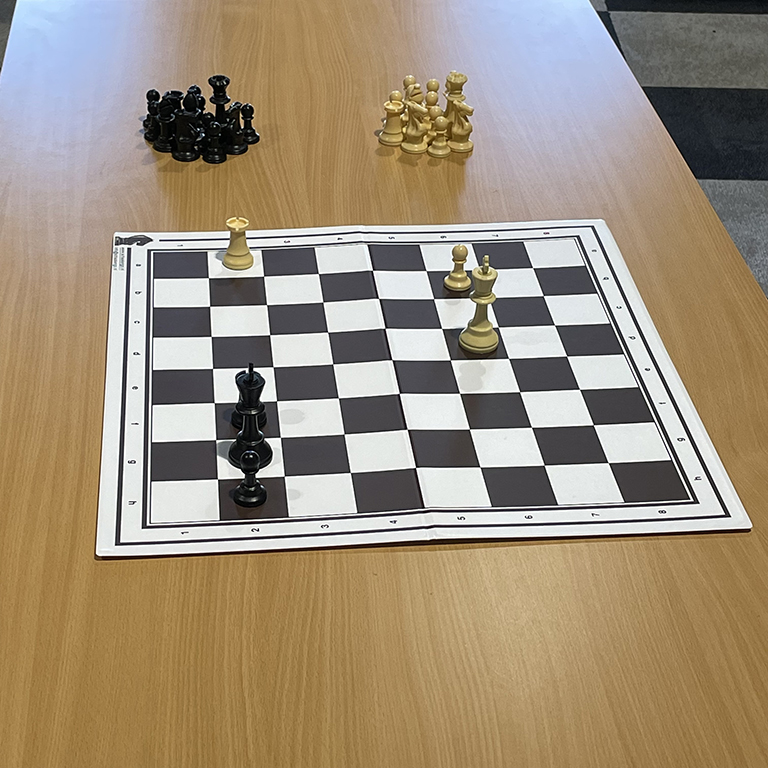}
  \end{subfigure}
  \hfill
  \begin{subfigure}[t]{.66\columnwidth}
    \centering
    \includegraphics[width=\linewidth]{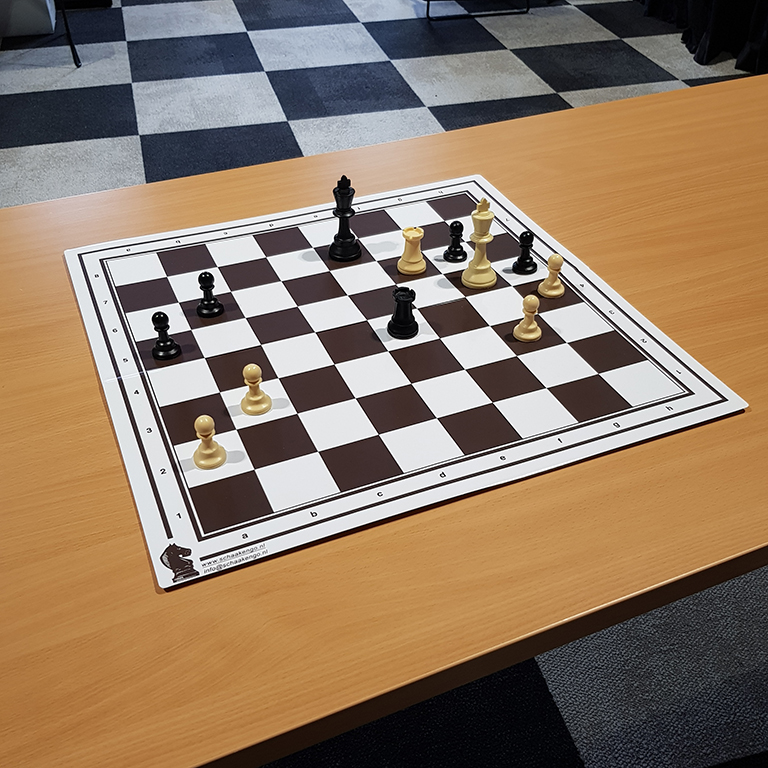}
  \end{subfigure}
  \hfill
  \caption{End-game (more than 75 moves) samples from ChessReD.} \label{apdx:fig:end-game}
\end{figure*}

\end{document}